\documentclass{article}
\usepackage{booktabs}   
\usepackage{multirow}  
\usepackage{amsmath}
\usepackage{PRIMEarxiv}
\usepackage{makecell}
\usepackage[utf8]{inputenc} 
\usepackage[T1]{fontenc}    
\usepackage{hyperref}       
\usepackage{url}            
\usepackage{booktabs}       
\usepackage{amsfonts}       
\usepackage{nicefrac}       
\usepackage{microtype}      
\usepackage{lipsum}
\usepackage{float}
\usepackage{enumitem}
\usepackage[LGR,T1]{fontenc}  
\usepackage{textalpha}         
\usepackage{fancyhdr}       
\usepackage{graphicx}       
\graphicspath{{media/}}     
\usepackage{xcolor}
\usepackage{authblk}
\usepackage{hyperref}
\begin{document}
  
\title{Multimodal Human-AI Synergy for Medical Imaging Quality Control: A Hybrid Intelligence Framework with Adaptive Dataset Curation and Closed-Loop Evaluation}
%
\author[1,†]{Zhi Qin}          
\author[1,†]{Qianhui Gui}      
\author[2,†]{Mouxiao Bian}     
\author[1]{Rui Wang}
\author[1]{Hong Ge}
\author[1]{Dandan Yao}
\author[1]{Ziying Sun}
\author[1]{Yuan Zhao}
\author[1]{Yu Zhang}
\author[1]{Hui Shi}
\author[3]{Dongdong Wang}
\author[1]{Chenxin Song}
\author[1]{Shenghong Ju}
\author[2]{Lihao Liu}
\author[2]{Junjun He}
\author[2,*]{Jie Xu}
\author[1,*]{Yuan-Cheng Wang}
 
\affil[1]{\textit{
    Department of Radiology, Zhongda Hospital \\
    Nurturing Center of Jiangsu Province for State Laboratory of AI Imaging \& Interventional Radiology \\
    School of Medicine, Southeast University\\
    Nanjing, China
}}
 
\affil[2]{\textit{
    Shanghai Artificial Intelligence Laboratory\\
    Shanghai, China
}}
 
\affil[3]{\textit{
    Department of Radiology\\
    The Fifth Clinical Medical College of Henan University of Chinese Medicine \\
    (Zhengzhou People's Hospital) \\
    Zhengzhou, China
}}
 
\footnotetext[1]{†These authors contributed equally.}
\footnotetext[2]{*Correspondence: 
Yuan-Cheng Wang (yuancheng\_wang@seu.edu.cn); 
Jie Xu (xujie@pjlab.org.cn)
}

\maketitle

\begin{abstract}

Medical imaging quality control (QC) is essential for accurate diagnosis, yet traditional QC methods remain labor-intensive and subjective. To address this challenge, in this study, we establish a standardized dataset and evaluation framework for medical imaging QC, systematically assessing large language models (LLMs) in image quality assessment and report standardization. Specifically, we first constructed and anonymized a dataset of 161 chest X-ray (CXR) radiographs and 219 CT reports for evaluation. Then, multiple LLMs, including Gemini 2.0-Flash, GPT-4o, and DeepSeek-R1, were evaluated based on recall, precision, and F1 score to detect technical errors and inconsistencies. Experimental results show that Gemini 2.0-Flash achieved a Macro F1 score of 90 in CXR tasks, demonstrating strong generalization but limited fine-grained performance. DeepSeek-R1 excelled in CT report auditing with a 62.23\% recall rate, outperforming other models. However, its distilled variants performed poorly, while InternLM2.5-7B-chat exhibited the highest additional discovery rate, indicating broader but less precise error detection. These findings highlight the potential of LLMs in medical imaging QC, with DeepSeek-R1 and Gemini 2.0-Flash demonstrating superior performance.

\end{abstract}


\section{Introduction}

Medical imaging reports and images are fundamental to clinical diagnosis, directly impacting diagnostic accuracy and treatment outcomes. The rapid advancement of medical imaging technologies has led to an exponential increase in imaging data, accompanied by a surge in the number of imaging reports\cite{mcdonald2015effects}. However, image quality can be compromised by various factors such as equipment settings, patient positioning, and procedural inconsistencies, leading to artifacts, insufficient resolution, or unclear anatomical visualization. Concurrently, standardizing imaging reports remains a challenge due to inconsistent terminology, inaccurate descriptions, and logical inconsistencies\cite{cai2016natural}\cite{treanor2021reporting}. These issues not only increase the risk of misinterpretation but also contribute to misdiagnoses or missed diagnoses, ultimately threatening patient safety.

Traditional quality control (QC) in medical imaging primarily relies on manual review, which is time-consuming, labor-intensive, and prone to subjective biases, making it increasingly inadequate for the growing demands of modern healthcare. In contrast, large language models (LLMs) have demonstrated remarkable progress in natural language processing, excelling in semantic understanding and generation\cite{liu2025generalist}\cite{qiu2024llm}\cite{goh2025gpt}. They have been widely applied in medical tasks such as text summarization, clinical decision support, and patient question-answering, showcasing significant practical value\cite{zhang2024pediabench}\cite{abrar2025empirical}\cite{thirunavukarasu2023large}. Recent studies also suggest that LLMs can enhance medical imaging report analysis by extracting disease features, generating structured reports\cite{adams2023leveraging}\cite{bosbach2024ability}, and assessing image quality\cite{gu2024using}\cite{fink2023potential}. By integrating imaging and textual information through multimodal learning, LLMs hold promise for automating image quality evaluation and report standardization, potentially assisting radiologists and reducing workload. Moreover, several models are now capable of generating imaging reports and performing preliminary image interpretation, supported by the establishment of various datasets and benchmarks\cite{kus2024medsegbench}\cite{metmer2024open}. However, despite these advancements, LLMs face two key challenges in medical imaging QC: \textbf{(1)} their performance varies significantly due to differences in model architecture, training data, and technical approaches. \textbf{(2)} the lack of standardized datasets annotated by experienced radiologist and mutimodal evaluation frameworks for imaging QC has hindered further optimization and practical application of LLMs in this domain.

To address the above two  challenges, this study establishes a standardized dataset and human-artificial intelligence(AI) synergy evaluation framework for medical imaging quality control. We systematically evaluate various LLMs, focusing on their potential in image quality assessment and report standardization. By integrating multimodal data and implementing a standardized evaluation framework, this research fills a critical gap in imaging quality control while providing a foundation for developing efficient and reliable intelligent QC tools. Ultimately, our work aims to advance the application of LLMs in medical imaging, offering innovative solutions to optimize clinical workflows and enhance healthcare efficiency and quality.

\section{Method}

The models and parameters utilized for evaluating the quality of chest X-ray radiographs and CT reports are configured as follows(Table 1):
\begin{table}[H]
  \caption{Medical Imaging Quality Control Model Comparison}
  \centering
  \begin{tabular}{lllll}
    \toprule
    Task Type & Model & Parameters & Context length & Open Source \\
    \midrule
    \multirow{5}{*}{Chest X-ray QC}& InternVL2.5-8B & 8B & 8K & Yes \\
    & Gemini 2.0 Flash & 540B & 1M & No \\
    & GPT-4o & About 200B & 4K & No \\
    & QVQ-72B-Preview & 72B & 32K & Yes \\
    & Qwen2.5-VL-72B-Instruct & 72B & 128K & Yes \\
    
    \midrule
    
    \multirow{13}{*}{CT Report QC}& InternLM2.5-7B-Chat & 7B & 32K & Yes \\
    & Deepseek-R1 & 671B & 128K & Yes \\
    & Deepseek-V3 & 671B & 128K & Yes \\
    & Gemini 1.5 Pro & 175B & 2M & No \\
    & Qwen2.5-32B-Instruct & 32B & 128K & Yes \\
    & Qwen2.5-72B-Instruct & 72B & 128K & Yes \\
    & Llama-3.1-405B-Instruct & 405B & 128K & Yes \\
    & Llama-3.3-70B-Instruct & 70B & 128K & Yes \\
    & DeepSeek-R1-Distill-Qwen-32B & 32B & 128K & Yes \\
    & DeepSeek-R1-Distill-Llama-70B & 70B & 128K & Yes \\
    & GPT-4o & About 200B & 128K & No \\
    & Gemma-2-9b-it & 9B & 8K & Yes \\
    & Claude 3.5 Sonnet & 175B & 200K & No \\
    
    \bottomrule
  \end{tabular}
  \label{tab:medical_models}
\end{table}

\section{Dataset}
\subsection{Data Collecting}
As illustrated in Figure 1, this study retrospectively collected high-resolution frontal  CXR radiographs and CT reports from the Radiology Department of Zhongda Hospital Southeast University, spanning the period from January 2023 to December 2024. To ensure data integrity and clinical relevance, the collection protocol was supervised by a panel of five board-certified radiologists with over 15 years of clinical experience each. Our dataset comprises 161 CXR radiographs and 219 structured CT reports extracted from the institutional Picture Archiving and Communication System (PACS), including non-audited cases for comprehensive analysis.
\begin{figure}
    \centering
    \includegraphics[width=1\linewidth]{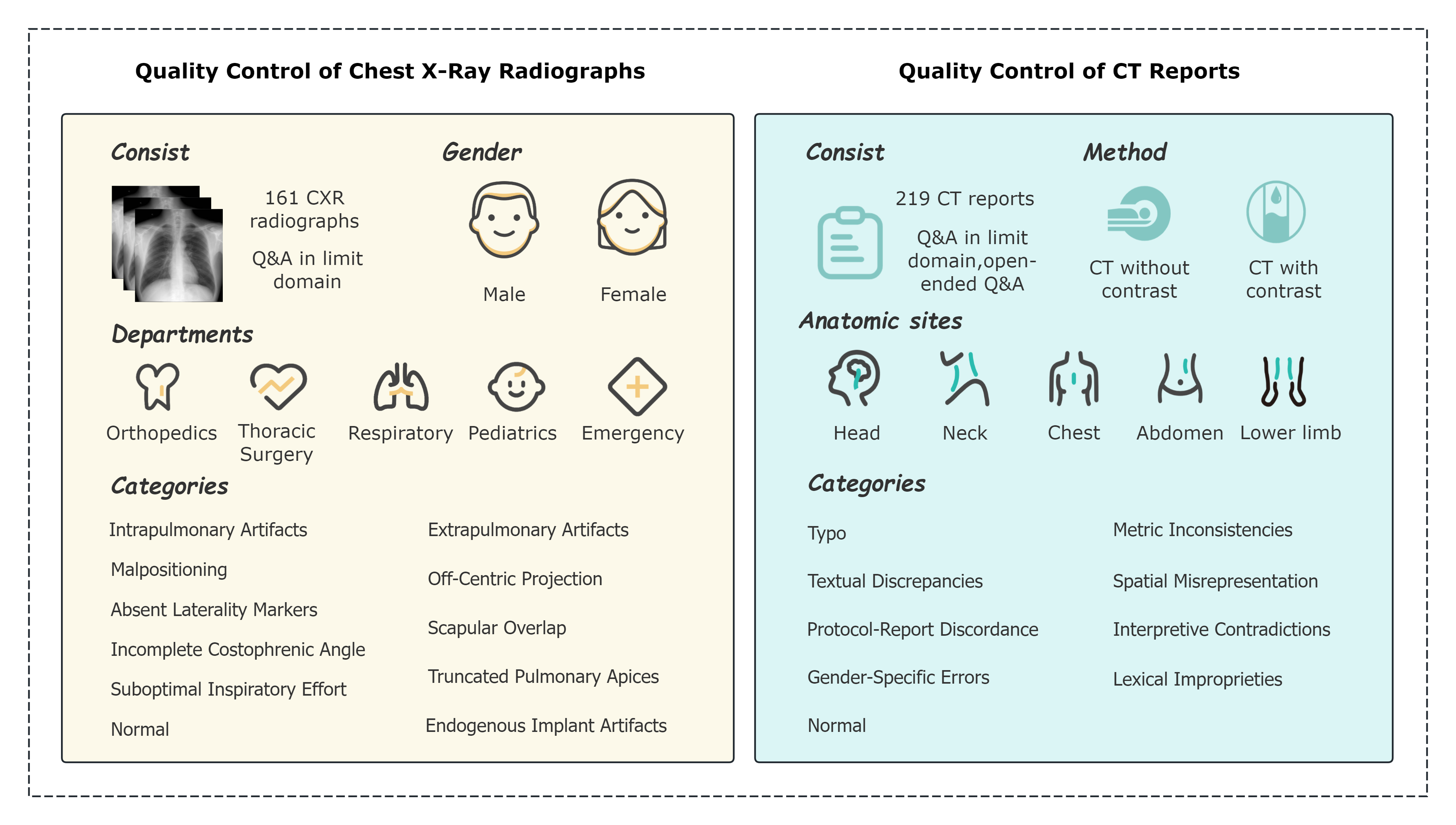}
    \caption{Medical Imaging Quality Control }
    \label{fig:enter-label}
\end{figure}
 All medical reports followed a standardized documentation framework consisting of two principal components: (1) objective imaging findings and (2) impressions. The data migration process employed hospital-approved encrypted transfer protocols with SHA-256 integrity verification to maintain data fidelity during transmission from PACS to local secure storage arrays.
In strict compliance with HIPAA standards\cite{hhs2008} and institutional review board (IRB) guidelines, we implemented a multi-stage de-identification pipeline. This included DICOM header anonymization using Python-GDCM libraries and report sanitization through a hybrid approach combining regular expression pattern matching with large language model (LLM)-based natural language processing (NLP) techniques. The anonymization process systematically removed all protected health information (PHI) while preserving critical clinical semantics through conditional random field (CRF) optimization prior to analytical processing.
\subsection{Quality Control Standards for CXR Radiographs}
The quality control standards for CXR radiographs were established through a systematic synthesis of 11 technical criteria, rigorously aligned with international guidelines from the American College of Radiology (ACR)\cite{acr2023tech}\cite{acr2023positioning}\cite{acr2023practice}\cite{acr2023radpeer}\cite{acr2023digital}, International Electrotechnical Commission (IEC 61223-3-5:2020)\cite{iec2020}, and Fleischner Society\cite{fleischner} (Table 2). These criteria address critical dimensions of radiographic adequacy, including artifact classification, patient positioning, and anatomical coverage. Intrapulmonary artifacts, defined as foreign objects or medical implants within aerated lung fields, were evaluated for their potential to obscure parenchymal details, with a threshold of 20\% visibility loss set as clinically significant based on multi-institutional consensus. Positioning accuracy mandates clavicular midline alignment within 5 mm of the tracheal air column and spinous process midline, validated through DICOM coordinate analysis (Python-GDCM). Projection geometry thresholds restrict craniocaudal deviations to ±5 mm from the T6 vertebral level and laterolateral deviations to ±10 mm from the spinal midline, ensuring consistent cardiothoracic ratio measurements.

Anatomical coverage criteria require complete visualization of costophrenic angles, defined by a sharp edge gradient of $\geq$3.0 Hounsfield Units (HU)/mm at the costo-diaphragmatic interface, and inclusion of$\geq$  95\% of the superior pulmonary margins. Scapular overlap within the lateral upper lung zones, manifested as linear hyperdensities parallel to the thoracic wall, was restricted to <30\% lung field coverage, a threshold derived from ACR Technical Standards to minimize diagnostic interference. Suboptimal inspiratory effort, quantified by the visualization of fewer than 10 posterior ribs or anterior rib interspace separation <3 mm, was linked to artifactual vascular equalization, as per Fleischner Society guidelines.

\begin{table}[htbp]
  \centering
  \caption{CXR Radiographs Quality Control Criteria}
  \label{tab:qc_criteria}
  \begin{tabular}{ll}
    \toprule
    \textbf{Error Type} & \textbf{Operational Definition} \\
    \midrule
    Intrapulmonary Artifacts & 
    Foreign objects/implants within aerated lung fields obscuring parenchymal details \\
    
    Extrapulmonary Artifacts & 
    High-density external objects overlapping mediastinal structures \\
    
    Malpositioning & 
    \makecell[l]{Clavicular asymmetry >5 mm from midline\\Tracheal air column misalignment} \\
    
    Off-Centric Projection & 
    \makecell[l]{Craniocaudal: >5 mm from T6 level\\Laterolateral: >10 mm from midline} \\
    
    Absent Laterality Markers & 
    Missing anatomical orientation markers ("R"/"L") \\
    
    Scapular Overlap & 
    >30\% lateral upper lung zone coverage by scapular shadows \\
    
    Incomplete Costophrenic Angle & 
    Unsharp costo-diaphragmatic interface or missing pleural recesses \\
    
    Truncated Pulmonary Apices & 
    >5\% exclusion of superior lung margins \\
    
    Suboptimal Inspiratory Effort & 
    \makecell[l]{<10 posterior ribs visualized\\Diaphragm cranial to 7th anterior rib\\Rib spacing <3 mm} \\
    
    Clavicular Malalignment & 
    \makecell[l]{Sternoclavicular asymmetry >5 mm\\Vertical displacement >3 mm} \\
    
    Endogenous Implant Artifacts & 
    \makecell[l]{Permanent foreign bodies causing non-removable interference:\\• Extrinsic: External metallic objects\\• Implant-related: Biomedical devices} \\
    \bottomrule
  \end{tabular}
\end{table}
\subsection{Quality Control Criteria for CT Reports}
Based on clinical practice experience and relevant literature, this study systematically categorizes errors in CT imaging reports into eight major types, encompassing issues ranging from terminology misuse to logical inconsistencies in diagnostic conclusions.\cite{vosshenrich2021revealing}\cite{ringler2015syntactic}\cite{gertz2024potential}\cite{iso2022} The specific classifications include: (1)Typo, (2) Metric Inconsistencies, (3) Textual Discrepancies, (4)Spatial Misrepresentation, (5) Protocol-Report Discordance, (6) Interpretive Contradictions, (7)Gender-Specific Errors, and (8) Lexical Improprieties. This classification framework aims to comprehensively cover potential quality issues in CT reports, providing a clear reference for subsequent model evaluation.

 To quantitatively assess the performance of large language models in detecting these error types, this study establishes a three-tier evaluation standard: (1) Model results align with the gold standard; (2) Model results do not fully align with the gold standard but improve report quality to some extent; and (3) Model results deviate from the gold standard, misinterpret the original report, and reduce report quality. This grading system is designed to evaluate model performance from both accuracy and practicality perspectives.

 To ensure objective and reliable evaluations, two senior radiologists (each with over three years of clinical experience) independently assessed all chest X-ray radiographs , CT reports, and model outputs. Scoring was based on predefined error classifications and criteria. Disagreements were resolved by a third senior physician with over 10 years of experience. Reviewers underwent standardized training to minimize subjective bias.

\subsection{Dataset Annotation}
To ensure compliance with ethical and regulatory standards, a multi-stage anonymization protocol was implemented. DICOM metadata underwent rigorous de-identification using Python-GDCM , stripping PHI such as patient identifiers while preserving critical technical parameters including tube voltage (kVp), tube current-time product (mAs), and source-to-image distance (SID). Textual reports were processed through a hybrid sanitization pipeline combining regular expression pattern matching for structured data (e.g., date formats, medical record numbers) and BioBERT-CRF models for context-aware PHI redaction in free-text narratives without compromising clinical semantics such as anatomical descriptors or pathological findings.

Validation of annotation fidelity followed a tripartite framework integrating automated, AI-assisted, and expert-driven methodologies. Initial automated quality checks verified DICOM tag consistency, including validation of anatomical laterality markers (DICOM tag 0020,9072) and cross-referencing Modality Worklist entries with protocol descriptions to detect discrepancies such as non-contrast CT scans erroneously labeled as contrast-enhanced studies, maintaining a tolerance threshold of <0.5\% for modality code mismatches. Subsequent AI-assisted analysis employed U-Net architectures for quantitative artifact assessment in chest radiographs, segmenting scapular overlap regions and enforcing a strict <30\% lung field coverage limit as defined by ACR Technical Standards. Simultaneously, ResNet-50 models enumerated posterior ribs to flag suboptimal inspiratory effort (visualization of <10 ribs) and measured anterior rib interspace separation (<3 mm), thresholds derived from Fleischner Society consensus guidelines.
\subsection{Prompt Design for Quality Control Tasks}
The dataset was structured to evaluate LLMs through clinically grounded query-response pairs that simulate real-world quality control workflows. For CXR photographs quality assessment, each case presented a high-resolution image with the standardized query:

\textit{\textit{"Based on the following frontal chest X-ray, determine whether any of the ten technical issues listed in Table 1 are present. If no issues exist, respond with 'No issues detected.' If issues are identified, list all applicable error types without elaboration."}}

This task required models to analyze positioning accuracy (e.g., clavicular alignment), anatomical coverage (e.g., truncated apices), and artifact interference (e.g., scapular overlap) based on predefined thresholds.
 For CT report quality control, cases included de-identified textual reports formatted as:
 
\textit{\textit{"Below is a CT report for a [age]-year-old [gender] patient who underwent [non-contrast/contrast-enhanced] [protocol name, e.g., Chest CT]. The report contains 'Findings' and 'Impression' sections. Evaluate for the eight error types in Table 2. If no errors exist, state 'No issues detected.' If errors are present, specify:}}
\textit{\textit{Error type(s)}}
\textit{\textit{Exact text segment requiring correction (quoted from Findings/Impression)}}
\textit{\textit{Proposed revision with rationale based on referenced standards.}}

The task architecture was engineered to reflect modality-specific clinical reasoning demands, with CXR radiographs quality control requiring spatial analysis of anatomical positioning (e.g., clavicular alignment deviations) and artifact distribution through image-to-error mapping, while CT report evaluation necessitated context-aware semantic understanding to detect inconsistencies between textual descriptors and imaging protocols. Actionable outputs were mandated to include localized error identification—specifying DICOM slice coordinates for CXR anomalies (e.g., off-centric projections >5 mm from T6) and exact text spans for CT report errors (e.g., "5 m lesion" in Findings)—paired with guideline-cited corrections referencing ACR Technical Standards or RSNA Reporting Initiative criteria\cite{rsna2023}. To rigorously assess model specificity, 5\% of cases were intentionally designed as error-free, challenging systems to minimize false positives when processing technically adequate CXR radiographs or semantically compliant CT reports. This unified structure enables precise evaluation of LLMs' capacity to operationalize QC standards in clinical workflows, balancing error detection sensitivity with diagnostic workflow compatibility through anatomically grounded image analysis and contextually aware text interpretation.

\section{Experiment}

\subsection{Ethical Statement}
This retrospective study received ethical approval , and the requirement for informed consent was waived, as no patient identifiers were used in any part of this retrospective study. The dataset solely comprised de-identified radiology reports in plain text and X-ray radiographs generated within our department. Neither the data nor the reports could be traced back to individual patients or radiologists, ensuring complete anonymity and compliance with ethical standards.
\subsection{Experimental Setting and Design}
In order to make our framework more scientific and rigorous, we designed our workflow based on the renowned medical large model evaluation platform—Medbench\cite{liu2024medbench} (Figure 2). At the same time, we will host a medical imaging specialty quality control competition on this platform, which can be accessed through the following link: \href{https://medbench.opencompass.org.cn/track}{https://medbench.opencompass.org.cn/track}.The validation framework employed an API-driven pipeline to evaluate 26 LLMs (18 LLMs, 8 MLLMs) under simulated clinical workflows. For CXR radiographs quality assessment, DICOM images were programmatically anonymized using Python-GDCM’s metadata scrubbing toolkit, converted to lossless PNGs, and embedded into structured prompts via RFC 4648 Base64 encoding. Multimodal API calls incorporated task-specific instructions aligned with ACR Technical Standards, ensuring standardized error definitions (e.g., clavicular misalignment >5 mm). CT report validation utilized JSON-formatted inputs preserving original report semantics.
\begin{figure}
    \centering
    \includegraphics[width=0.8\linewidth]{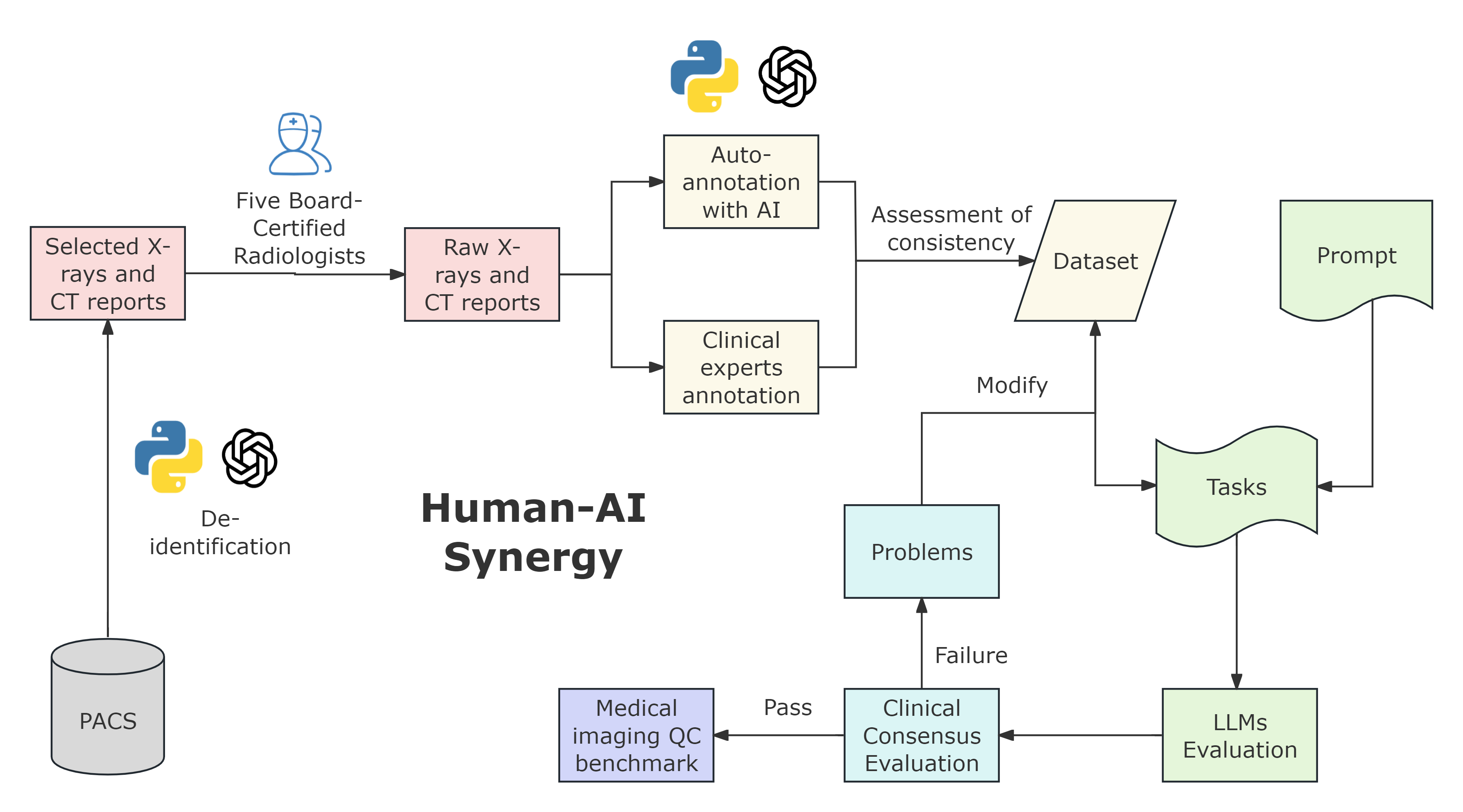}
    \caption{Model Validation Framework}
    \label{fig:enter-label}
\end{figure}

\subsection{Evaluation Metrics}
\subsubsection{Chest X-Ray Quality Control}
The performance of quality control systems was rigorously evaluated through modality-specific metrics aligned with clinical validation protocols.
Models were assessed using  \textbf{F1} \textbf{score} for recognition accuracy:
\begin{enumerate}[label=\arabic*), nosep, leftmargin=*, align=parleft] 
  \item \textbf{True Positives (TP)}: required exact identification of all technical errors listed in Table~2.
  \item \textbf{False Positives (FP)}: incurred full penalty on gold-standard negative studies.
  \item \textbf{False Negatives (FN)}: represented missed identification of any technical errors or errors present in flawless cases.
\end{enumerate}
The Micro-F1 score, which combines precision and recall into a single metric, was used to evaluate the model performance across all classes, weighted by the number of instances in each class.
$$
\text{Micro-F1} = \frac{2 \times \text{Precision} \times \text{Recall}}{\text{Precision} + \text{Recall}}
$$
\textit{\textit{Where:}}
\textit{\textit{Precision: The ratio of correctly identified errors to all cases predicted as errors.}}
\textit{\textit{Recall: The ratio of correctly identified errors to all actual errors present in the data.}}
The Macro-F1 methodology was specifically adopted to address categorical detection parity in multi-class quality assessment scenarios. For each distinct technical error category k ($1 \leq k \leq K$, where K = 15 as per Table 1), class-specific confusion matrices were constructed independently. This approach deliberately decouples error type prevalence from performance evaluation, counteracting the dataset's inherent imbalance where certain artifacts (e.g., rotation errors) occur 23× more frequently than critical low-prevalence errors (e.g., misaligned side markers). The per-class F1 computation ensures that models achieving high performance through dominant class overfitting are penalized, as demonstrated by the metric's sensitivity to underperformance in any single category:
\begin{equation}
\text{Macro-F1} = \frac{1}{K} \sum_{k=1}^{K} \left( \frac{2 \cdot TP_k}{2 \cdot TP_k + FP_k + FN_k} \right)
\end{equation}

This formulation assigns equivalent weight to all error classes, effectively modeling the clinical prioritization principle that missing a single critical error (e.g., incorrect laterality markers) carries equivalent consequence to missing multiple common artifacts. Validation included comparison against alternative frameworks (weighted-F1, Cohen's κ), with Macro-F1 selected for its stricter enforcement of cross-class performance consistency – a prerequisite for regulatory-compliant quality assurance systems where no error type can be systematically underdetected.
\subsubsection{CT Reports  Quality Control}
\textbf{\textbf{Micro-F1 }}\textbf{\textbf{Score}}
\textbf{1) }\textbf{\textbf{True Positives (TP)}} represents model results align with the gold standard.
\textbf{2) }\textbf{\textbf{False Negatives (FN)}} represents model results deviate from the gold standard, misinterpret the original report, and reduce report quality.

$$
\text{Micro-F1} = \frac{2 \times \text{Precision} \times \text{Recall}}{\text{Precision} + \text{Recall}}
$$
\textit{\textit{Where:}}
\textit{\textit{Precision: The ratio of correctly identified errors to all cases predicted as errors.}}
\textit{\textit{Recall: The ratio of correctly identified errors to all actual errors present in the data.}}
This emphasizes strict alignment with gold-standard annotations, where FP penalizes overconfident corrections and FN penalizes critical oversight errors.
\subsubsection{Statistical Analysis}
All analyses were performed using the 3.8.0 version of Python, the formatting of model output results and data analysis has been implemented. All charts were completed using the ggplot2 package in R.

\section{Result}
\subsection{Performance Evaluation of Multimodal Large Language Models in Chest X-ray Quality Control }

We evaluated the performance of several multimodal large language models (MLLMs) on CXR radiographs quality control tasks using both Micro F1 and Macro F1 scores. The models assessed included Gemini 2.0-Flash, GPT-4o, InternVL2-8B, QVQ-72B Preview, and Qwen2.5-VL-72B-Instruct. To facilitate a clearer comparison of performance differences among the models, we normalized the F1 scores by scaling the highest-performing model's score to 90 and proportionally adjusting the scores of the remaining models to derive normalized F1 scores.
\begin{figure}[H]
    \centering
    \includegraphics[width=0.5\linewidth]{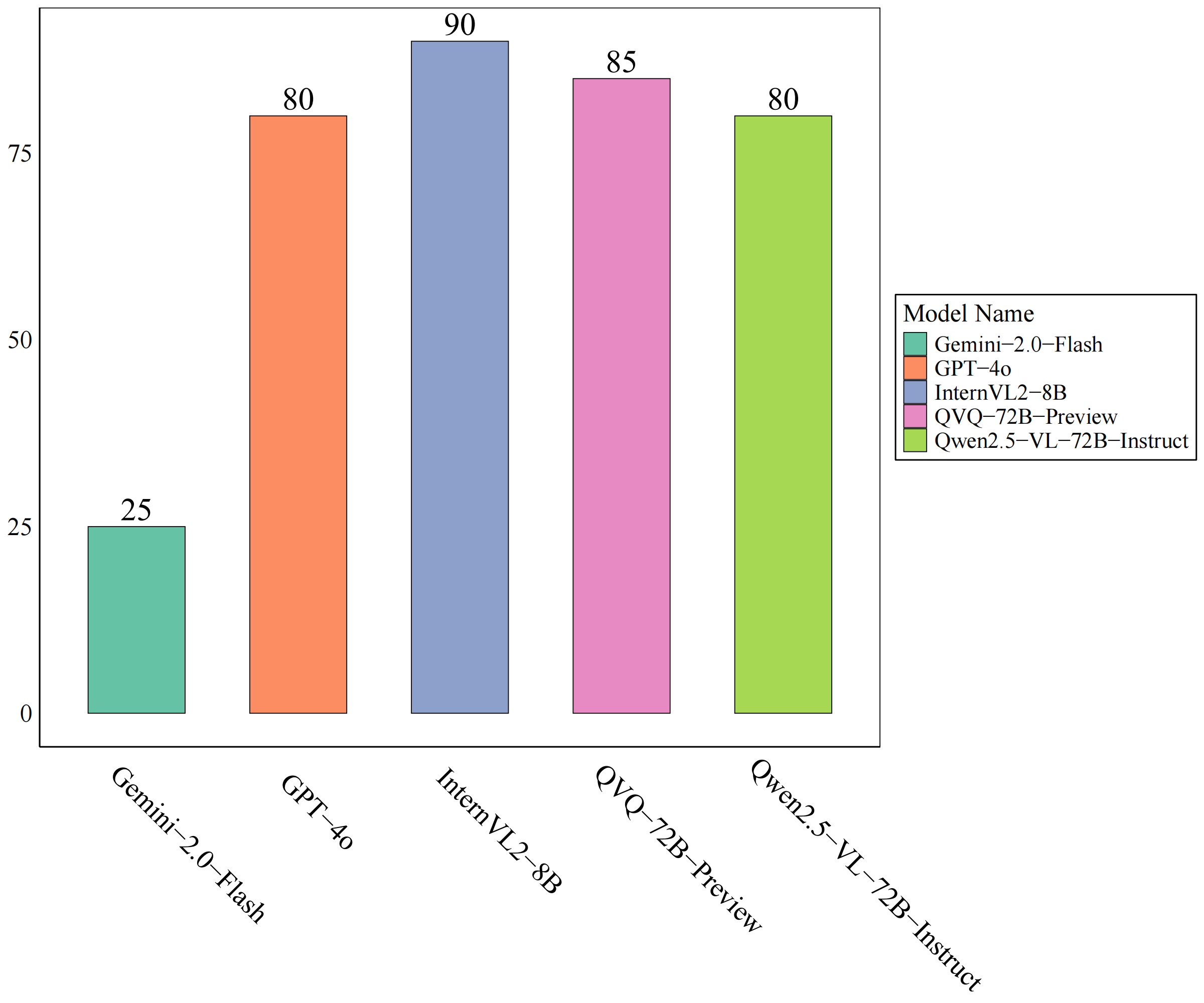}
    \caption{Normalized Micro F1 Score of MLLMs in CXR QC }
    \label{fig:enter-label}
\end{figure}
 As illustrated in Figure 3, Gemini 2.0-Flash achieved the lowest Micro F1 score of 25, significantly underperforming compared to other models. GPT-4o followed closely with a score of 80, demonstrating robust performance in CXR quality control tasks. InternVL2-8B and QVQ-72B Preview scored 90 and 85, respectively, indicating their strong capability in fine-grained task processing. Qwen2.5-VL-72B-Instruct, similar to InternVL2-8B, achieved a notable Micro F1 score of 80, suggesting that its instruction-based learning approach is competitive with other leading models.

\begin{figure}[H]
    \centering
    \includegraphics[width=0.5\linewidth]{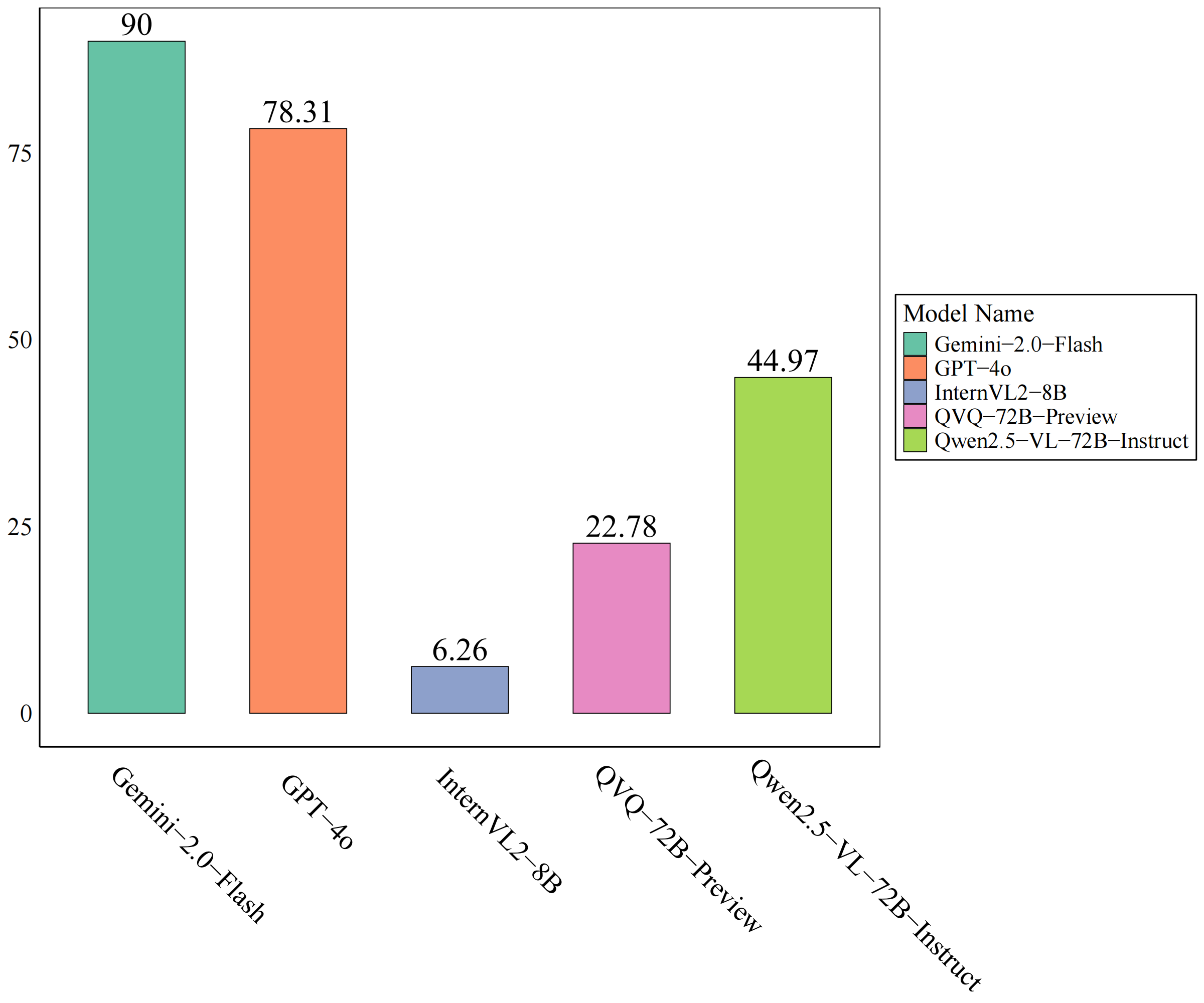}
    \caption{Normalized Macro F1 Score of MLLMs in CXR QC }
    \label{fig:enter-label}
\end{figure}
 The Macro F1 scores, presented in Figure 4, revealed that Gemini 2.0-Flash maintained a significant lead with a score of 90, highlighting its superior generalization ability across diverse CXR quality control categories. GPT-4o achieved a score of 78.31, demonstrating solid performance but falling short of Gemini 2.0-Flash. In contrast, InternVL2-8B scored only 6.26, indicating limited effectiveness in handling complex and varied CXR cases. QVQ-72B Preview and Qwen2.5-VL-72B-Instruct scored 22.78 and 44.97, respectively, with the latter outperforming other instruction-based models.

 The results demonstrate that Gemini 2.0-Flash achieves outstanding performance in both Micro F1 and Macro F1 scores, exhibiting high precision and recall across the dataset, making it an ideal candidate for widespread application in radiology workflows. Its superior generalization ability in diverse CXR radiographs quality control tasks suggests that the model can significantly reduce radiologists' workload by automating the detection of common imaging issues, such as mispositioning or artifacts, while maintaining diagnostic accuracy.
\begin{table}[htbp]
  \centering
  \caption{Model Performance Comparison by Error Type}
  \label{tab:error_comparison}
  \begin{tabular}{>{\raggedright\arraybackslash}p{0.26\linewidth}>{\centering\arraybackslash}p{0.15\linewidth}>{\centering\arraybackslash}p{0.12\linewidth}>{\centering\arraybackslash}p{0.12\linewidth}>{\centering\arraybackslash}p{0.1\linewidth}>{\centering\arraybackslash}p{0.12\linewidth}}
    \toprule
    \textbf{Error Type} & \textbf{Qwen2.5-VL-72B-Instruct}& \textbf{QVQ-72B-Preview}& \textbf{InternVL2.5-8B}& \textbf{GPT-4o} & \textbf{Gemini 2.0 Flash}\\
    \midrule
    Off-Centric Projection      & 0.00 & 0.00 & 0.00 & 0.08 & 0.16 \\
    Malpositioning              & 0.00 & 0.00 & 0.00 & 0.37 & 0.16 \\
    Scapular Overlap            & 0.28 & 0.04 & 0.00 & 0.31 & 0.41 \\
    Clavicular Malalignment     & 0.16 & 0.00 & 0.00 & 0.26 & 0.16 \\
    Error-free                  & 0.33 & 0.21 & 0.20 & 0.27 & 0.24 \\
    Incomplete Costophrenic     & 0.08 & 0.00 & 0.00 & 0.09 & 0.06 \\
    Truncated Pulmonary Apices  & 0.00 & 0.00 & 0.00 & 0.00 & 0.10 \\
    Extrapulmonary Artifacts    & 0.17 & 0.10 & 0.00 & 0.44 & 0.73 \\
    Absent Laterality Markers   & 0.12 & 0.00 & 0.00 & 0.07 & 0.00 \\
    Endogenous Implant Artifacts& 0.16 & 0.15 & 0.00 & 0.27 & 0.25 \\
    Intrapulmonary Artifacts    & 0.03 & 0.16 & 0.00 & 0.10 & 0.13 \\
    Suboptimal Inspiratory Effort & 0.10 & 0.07 & 0.00 & 0.25 & 0.50 \\
    \bottomrule
  \end{tabular}
\end{table}
\subsection{Performance Evaluation of Multimodal Large Language Models in CT Report Quality Control}
We assessed the performance of various LLMs in meeting medical question-answering standards, quality control, and task difficulty compliance. The evaluation included metrics such as compliance rate, F1 scores, and error categorization.
\begin{figure}
    \centering
    \includegraphics[width=0.8\linewidth]{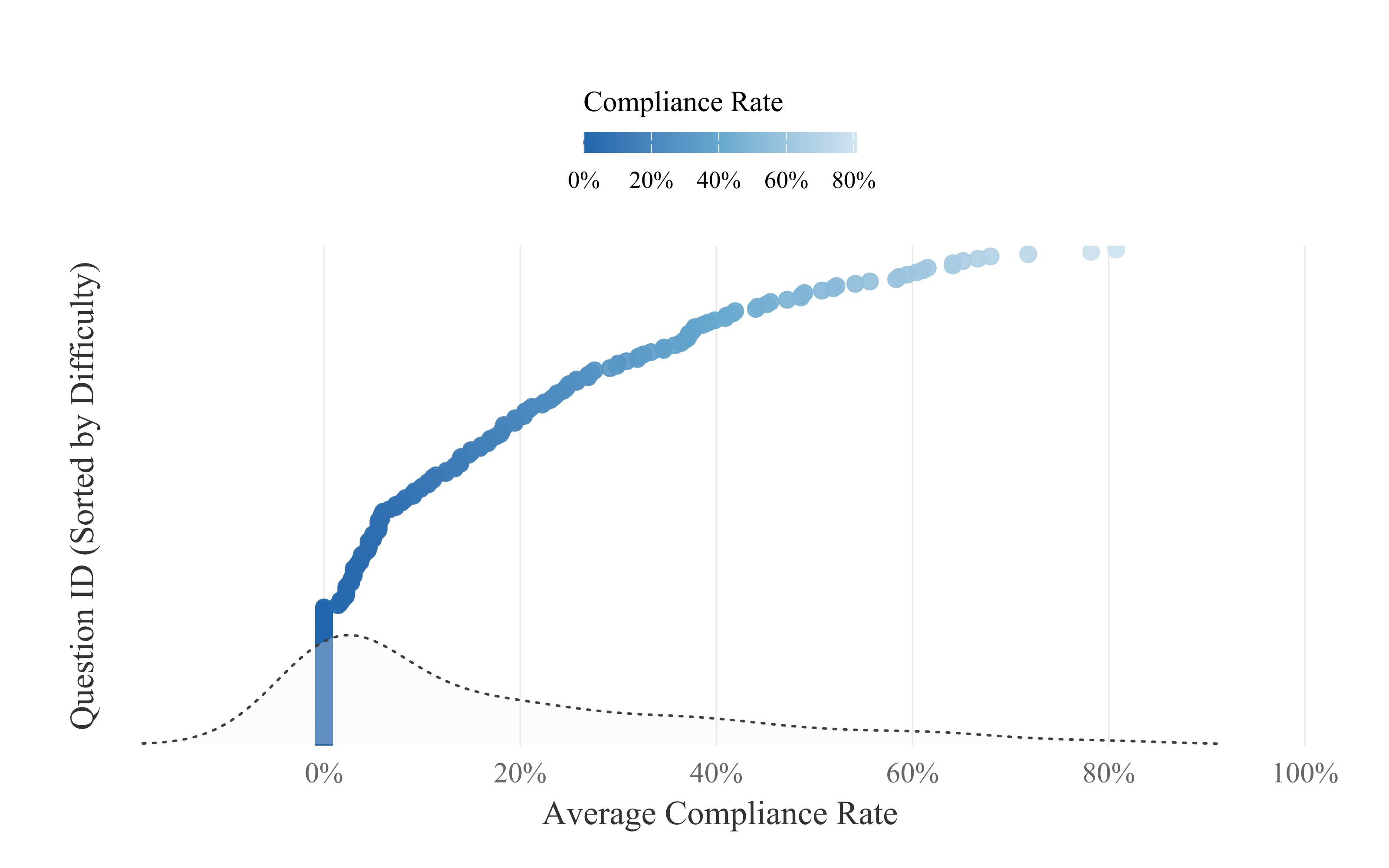}
    
    \caption{Distribution of Difficulty}
    \label{fig:enter-label}
\end{figure}
 First, the number of evaluation ratings for each question across all models was counted, and the correct response rate was calculated as the ratio of the number of times an answer was rated as "1" by human evaluators to the total number of responses for that question. The difficulty index was defined as 1 minus correct response rate, with values normalized to a range of 0 to 1. The questions were then divided into five difficulty levels using an equidistant binning method.As shown in Figure 5, the difficulty distribution of questions was analyzed across multiple models. The results indicate a gradual increase in difficulty as the compliance rate rises. Questions were sorted by difficulty, revealing that higher compliance rates were associated with lower difficulty levels, while more challenging questions exhibited lower compliance rates. This suggests that simpler questions were consistently answered correctly across models, whereas complex questions resulted in varying levels of success. The overall trend follows a characteristic learning curve, where model performance improves as they adapt to the difficulty distribution.

 Our analysis of model responses revealed that the probability of all answers being incorrect was approximately 40.15\%, while the probability of all answers being fully correct was only 13.3\%. Among the 219 questions, 61 had no model achieving complete accuracy, and questions with an average correct response rate below 0.2 accounted for 73.52\%. Given the absence of dedicated datasets for CT report quality control, this task presents a significant challenge for existing models, making the dataset particularly difficult for current state-of-the-art models.
\begin{figure}
    \centering
    \includegraphics[width=0.8\linewidth]{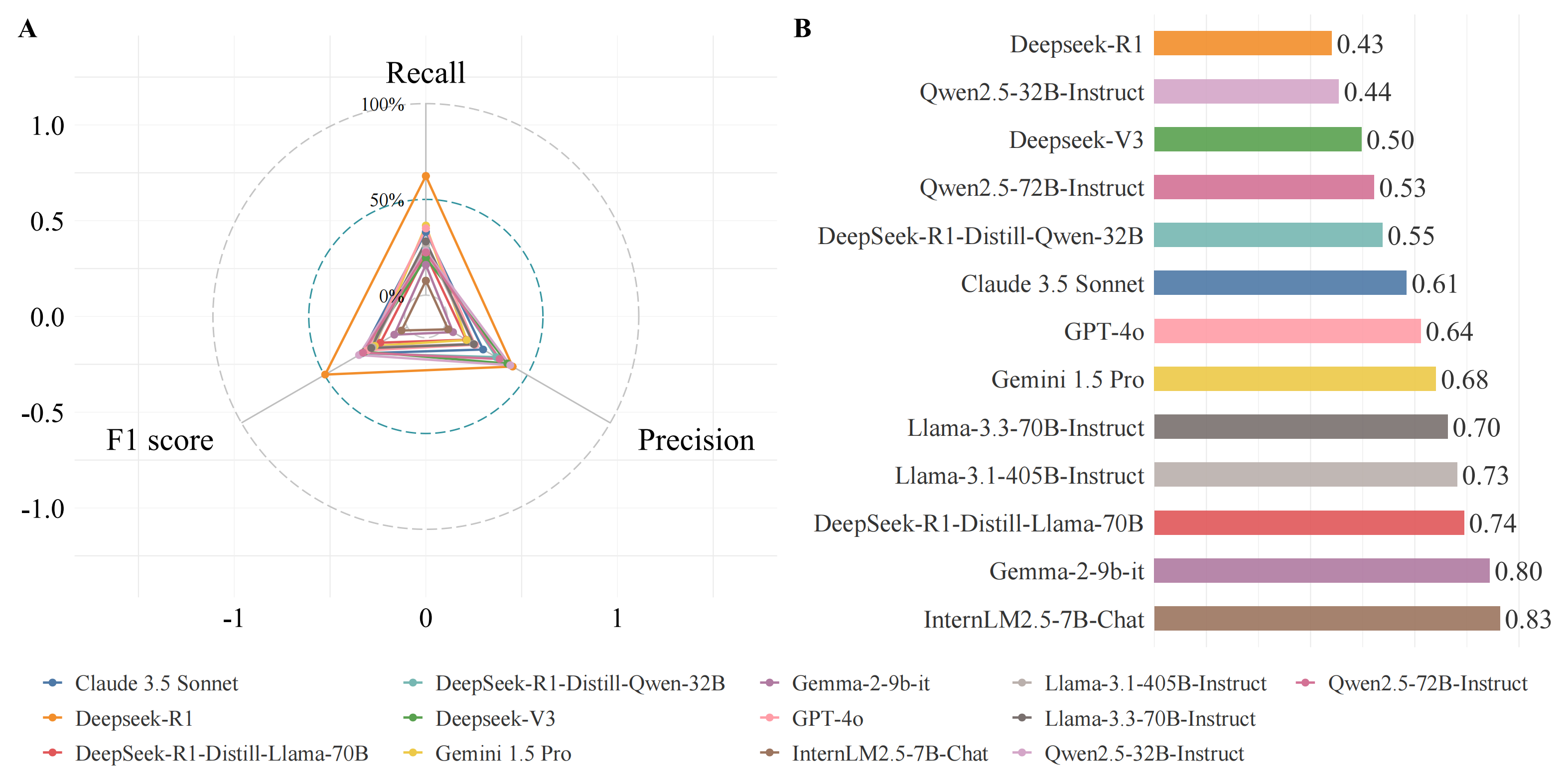}
    \caption{Performance Evaluation of LLMs in CT Report QC}
    \label{fig:enter-label}
\end{figure}
However, As clearly illustrated in Figure 6, DeepSeek-R1 outperformed other models in recall, precision, and F1 score for CT report quality control tasks, achieving a recall rate of 62.23\%, significantly higher than other models. The distilled variants of DeepSeek-R1, namely DeepSeek-R1-Distill-Qwen-32B and DeepSeek-R1-Distill-Llama-70B, demonstrated notably inferior performance compared to the original model. Interestingly, despite the CT reports being in Chinese, Gemini and GPT achieved F1 scores second only to DeepSeek-R1, surpassing DeepSeek-V3 and the Qwen series. However, DeepSeek-R1 exhibited the lowest additional discovery rate, while InternLM2.5-7B-Chat, despite having the lowest F1 score, achieved the highest additional discovery rate.
\begin{figure}
    \centering
    \includegraphics[width=1\linewidth]{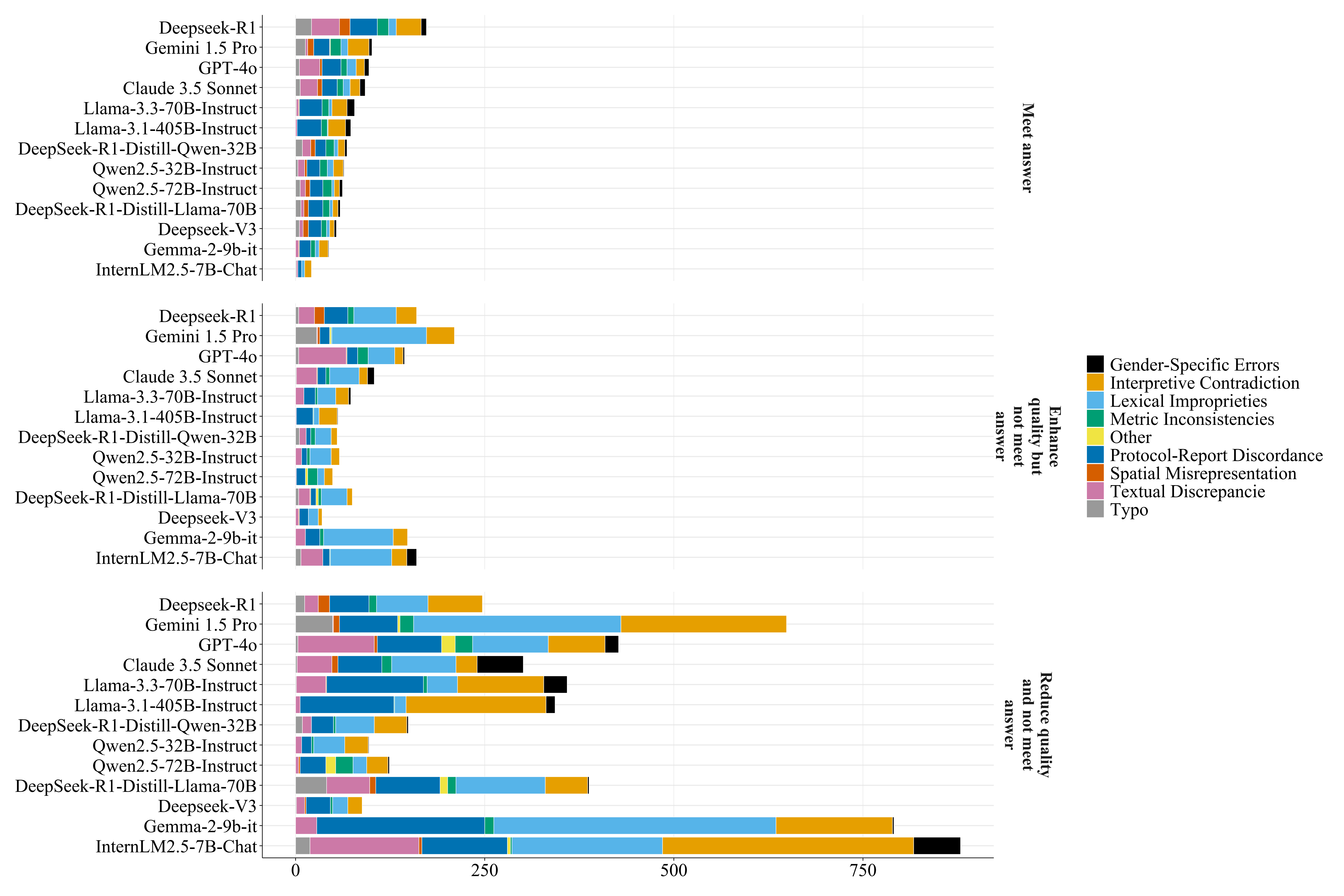}
    \caption{Performance Evaluation of LLMs in CT Report QC by Error Type.}
    \label{fig:enter-label}
\end{figure}
DeepSeek-R1 and Gemini 1.5 Pro excelled in providing correct answers ("Meet answer"), with fewer errors in other categories(Figure 7). Conversely, Qwen2.5-72B-Instruct and DeepSeek-R1-Distill-Qwen-32B showed improvement in the "Enhance quality but not meet answer" category, suggesting their responses were often high-quality but incomplete. InternLM2.5-7B-Chat and Gemma-2-9B-it contributed significantly to the "Reduce quality and not meet answer" category, frequently providing suboptimal or incomplete answers.

\section{Discussion}
\subsection{The advantage of our work}
This study represents the first effort to establish a standardized multimodal dataset and evaluation framework for medical imaging quality control tasks, and to comprehensively assess the performance of various MLLMs without task-specific training in the context of medical imaging quality control and report review. The experimental results reveal that while the majority of models performed at levels close to random chance in detecting irregularities in images and reports, certain models demonstrated notable potential\cite{lexa2021artificial}\cite{marcovici2009re}. Specifically, Gemini 2.0-Flash exhibited superior performance in CXR radiographs quality control tasks, while DeepSeek-R1 achieved exceptional results in CT reports review tasks. These findings not only validate the theoretical feasibility of leveraging large language models for imaging quality control tasks but also provide critical technical insights and optimization directions for future research endeavors.

 This study pioneers a standardized dataset for medical imaging quality control, defining multiple error types based on domestic and international guidelines, covering core quality control aspects with balanced proportions\cite{vosshenrich2021revealing}\cite{ringler2015syntactic}\cite{coppola2021human}. Derived from real imaging reports, the dataset comprehensively addresses key quality control issues. Results show that LLMs can detect both obvious errors and subtle imperfections, improving report quality, which reflects the dataset's comprehensiveness.
\subsection{Analysis for Performance of LLMs}
 In CXR radiographs quality control tasks, Gemini 2.0-Flash achieved a remarkable Macro F1 score of 90, excelling in generalization across diverse QC categories, likely due to its advanced multimodal framework and extensive pre-training. However, its low Micro F1 score (25) indicates limitations in fine-grained task processing, possibly due to insufficient domain-specific fine-tuning or challenges in handling subtle artifacts. GPT-4o and Qwen2.5-VL-72B-Instruct performed well in Micro F1 but showed limited generalization in Macro F1, while InternVL2-8B, despite strong Micro F1, scored poorly in Macro F1, highlighting the need to balance fine-grained performance and generalization\cite{fast2024autonomous}\cite{riccardi2024two}\cite{singhal2025toward}. In CT report quality control, DeepSeek-R1 demonstrated exceptional performance with a 62.23\% recall rate, attributed to its robust multimodal architecture and specialized fine-tuning, enabling effective detection of mismatched descriptions and logical errors\cite{fast2024autonomous}.In contrast, its distilled variants (e.g., DeepSeek-R1-Distill-Qwen-32B) exhibited inferior performance, underscoring the importance of maintaining model complexity and domain-specific knowledge in medical imaging QC tasks.

 Error categorization analysis revealed that DeepSeek-R1 and Gemini 1.5 Pro excelled in providing correct answers ("Meet answer"), with fewer errors in other categories, indicating their suitability for tasks requiring high accuracy and consistency. However, DeepSeek-R1 exhibited the lowest additional discovery rate, suggesting limitations in identifying novel or rare errors. In contrast, InternLM2.5-7B-Chat, despite its low F1 score, achieved the highest additional discovery rate, indicating a broader but less precise error detection capability. This trade-off between precision and discovery highlights the need for models that can balance both aspects in medical QC tasks.
\subsection{Limitations}
 Despite these achievements, the study has several limitations. First, the limited dataset size may restrict the models' generalization ability, particularly in handling rare issues. Second, the lack of transparency in the models' decision-making processes may affect clinicians' trust in the results. Additionally, the data primarily originated from a single institution, potentially introducing regional or device biases, which limits the models' generalizability. Furthermore, the current dataset is limited to Chinese-language reports, which may hinder its applicability in global clinical settings. To address these limitations, future research should focus on expanding dataset size and diversity, incorporating more imaging modalities (e.g., MRI, ultrasound) and a broader range of QC issues, including English-language reports and multimodal reports that integrate text, images, and other data formats. Developing domain-specific datasets and evaluation frameworks to enhance model performance is also crucial. Techniques such as instruction prompt tuning, RLHF, and adversarial training can further improve model accuracy and generalization\cite{thirunavukarasu2023large}\cite{singhal2023large}. Additionally, exploring explainable AI techniques to develop models with transparent decision-making processes could increase their clinical acceptability. Hybrid approaches that combine the strengths of different models (e.g., DeepSeek-R1's precision and InternLM2.5-7B-Chat's discovery capabilities) may also yield more robust solutions. Finally, multicenter studies should be conducted to validate model performance across different institutions, device conditions, and language settings, facilitating their practical application in clinical environments. Moreover, the current error classification and evaluation remain somewhat subjective, relying heavily on expert annotations. Future work should explore more objective quality control metrics, potentially leveraging automated scoring systems or quantitative measures to reduce subjectivity and enhance the reliability of model assessments.

 In conclusion, this study developed a dataset for CXR radiographs and CT report quality control, validating multiple LLMs. While most models performed suboptimally, the database effectively highlighted differences among them, revealing significant potential for distinguishing their detection capabilities. Gemini 2.0-Flash and DeepSeek-R1 excelled in CXR radiographs and CT tasks, respectively. Future work will expand the dataset to include more language modalities and error types, establishing a standardized benchmark to promote LLMs' clinical application and improve diagnostic efficiency.

\bibliography{references.bib} 

\begin{thebibliography}{10}
\providecommand{\url}[1]{#1}
\csname url@samestyle\endcsname
\providecommand{\newblock}{\relax}
\providecommand{\bibinfo}[2]{#2}
\providecommand{\BIBentrySTDinterwordspacing}{\spaceskip=0pt\relax}
\providecommand{\BIBentryALTinterwordstretchfactor}{4}
\providecommand{\BIBentryALTinterwordspacing}{\spaceskip=\fontdimen2\font plus
\BIBentryALTinterwordstretchfactor\fontdimen3\font minus \fontdimen4\font\relax}
\providecommand{\BIBforeignlanguage}[2]{{%
\expandafter\ifx\csname l@#1\endcsname\relax
\typeout{** WARNING: IEEEtran.bst: No hyphenation pattern has been}%
\typeout{** loaded for the language `#1'. Using the pattern for}%
\typeout{** the default language instead.}%
\else
\language=\csname l@#1\endcsname
\fi
#2}}
\providecommand{\BIBdecl}{\relax}
\BIBdecl

\bibitem{mcdonald2015effects}
R.~J. Mcdonald, K.~Schwartz, L.~Eckel \emph{et~al.}, ``The effects of changes in utilization and technological advancements of cross-sectional imaging on radiologist workload,'' \emph{Academic Radiology}, vol.~22, no.~9, pp. 1191--1198, 2015.

\bibitem{cai2016natural}
T.~Cai, A.~A. Giannopoulos, S.~Yu \emph{et~al.}, ``Natural language processing technologies in radiology research and clinical applications,'' \emph{Radiographics}, vol.~36, no.~1, pp. 176--191, 2016.

\bibitem{treanor2021reporting}
L.~M. Treanor, R.~Frank, A.~Atyani \emph{et~al.}, ``Reporting bias in imaging diagnostic test accuracy studies,'' \emph{American Journal of Roentgenology}, vol. 216, no.~1, pp. 225--232, 2021.

\bibitem{liu2025generalist}
X.~Liu, H.~Liu, G.~Yang \emph{et~al.}, ``A generalist medical language model for disease diagnosis assistance,'' \emph{Nature Medicine}, 2025.

\bibitem{qiu2024llm}
J.~Qiu, K.~Lam, G.~Li \emph{et~al.}, ``Llm-based agentic systems in medicine and healthcare,'' \emph{Nature Machine Intelligence}, vol.~6, no.~12, pp. 1418--1420, 2024.

\bibitem{goh2025gpt}
E.~Goh, R.~Gallo, E.~Strong \emph{et~al.}, ``Gpt-4 assistance for improvement of physician performance on patient care tasks: a randomized controlled trial,'' \emph{Nature Medicine}, 2025.

\bibitem{zhang2024pediabench}
Q.~Zhang, P.~Chen, J.~Li \emph{et~al.}, ``Pediabench: A comprehensive chinese pediatric dataset for benchmarking large language models,'' 2024.

\bibitem{abrar2025empirical}
M.~Abrar, Y.~Sermet, and I.~Demir, ``An empirical evaluation of large language models on consumer health questions,'' \emph{BioMedInformatics}, vol.~5, no.~1, p.~12, 2025.

\bibitem{thirunavukarasu2023large}
A.~Thirunavukarasu, D.~Ting, K.~Elangovan \emph{et~al.}, ``Large language models in medicine,'' \emph{Nature Medicine}, vol.~29, no.~8, pp. 1930--1940, 2023.

\bibitem{adams2023leveraging}
L.~Adams, D.~Truhn, F.~Busch \emph{et~al.}, ``Leveraging gpt-4 for post hoc transformation of free-text radiology reports into structured reporting: A multilingual feasibility study,'' \emph{Radiology}, vol. 307, no.~4, p. e230725, 2023.

\bibitem{bosbach2024ability}
W.~Bosbach, J.~Senge, B.~Nemeth \emph{et~al.}, ``Ability of chatgpt to generate competent radiology reports for distal radius fracture by use of rsna template items and integrated ao classifier,'' \emph{Current Problems in Diagnostic Radiology}, vol.~53, no.~1, pp. 102--110, 2024.

\bibitem{gu2024using}
K.~Gu, J.~Lee, J.~Shin \emph{et~al.}, ``Using gpt-4 for li-rads feature extraction and categorization with multilingual free-text reports,'' \emph{Liver International}, vol.~44, no.~7, pp. 1578--1587, 2024.

\bibitem{fink2023potential}
M.~Fink, A.~Bischoff, C.~Fink \emph{et~al.}, ``Potential of chatgpt and gpt-4 for data mining of free-text ct reports on lung cancer,'' \emph{Radiology}, vol. 308, no.~3, p. e231362, 2023.

\bibitem{kus2024medsegbench}
Z.~Kuş and M.~Aydin, ``Medsegbench: A comprehensive benchmark for medical image segmentation in diverse data modalities,'' \emph{Scientific Data}, vol.~11, no.~1, p. 1283, 2024.

\bibitem{metmer2024open}
H.~Metmer and X.~Yang, ``An open chest x-ray dataset with benchmarks for automatic radiology report generation in french,'' \emph{Neurocomputing}, vol. 609, p. 128478, 2024.

\bibitem{hhs2008}
\emph{Personal health records and the hipaa privacy rule}, U.S. Department of Health and Human Services, Washington, DC, 2008.

\bibitem{acr2023tech}
{American College of Radiology}, \emph{ACR Technical Standards for Diagnostic Radiology}, 2023.

\bibitem{acr2023positioning}
------, \emph{ACR Positioning Manual}, 2023.

\bibitem{acr2023practice}
------, \emph{ACR Practice Parameter for Imaging}, 2023.

\bibitem{acr2023radpeer}
------, \emph{{RADPEER{\textregistered} Scoring System}}, 2023.

\bibitem{acr2023digital}
{American College of Radiology}, {American Association of Physicists in Medicine}, and {Society for Imaging Informatics in Medicine}, \emph{Practice Guideline for Digital Radiography}, 2023.

\bibitem{iec2020}
\emph{Medical Imaging—Acceptance Testing and Quality Control—Part 3-5: Quality Control of X-ray Equipment for Diagnostic Radiology}, International Electrotechnical Commission Std. IEC 61\,223-3-5:2020, 2020.

\bibitem{fleischner}
{Fleischner Society}, \emph{Guidelines for the Management of Pulmonary Nodules}, n.d.

\bibitem{vosshenrich2021revealing}
J.~Vosshenrich, I.~Nesic, J.~Cyriac \emph{et~al.}, ``Revealing the most common reporting errors through data mining of the report proofreading process,'' \emph{European Radiology}, vol.~31, no.~4, pp. 2115--2125, 2021.

\bibitem{ringler2015syntactic}
M.~Ringler, B.~Goss, and B.~Bartholmai, ``Syntactic and semantic errors in radiology reports associated with speech recognition software,'' \emph{Studies in Health Technology and Informatics}, vol. 216, p. 922, 2015.

\bibitem{gertz2024potential}
R.~Gertz, T.~Dratsch, A.~Bunck \emph{et~al.}, ``Potential of gpt-4 for detecting errors in radiology reports: Implications for reporting accuracy,'' \emph{Radiology}, vol. 311, no.~1, p. e232714, 2024.

\bibitem{iso2022}
\emph{Quantities \& Units}, International Organization for Standardization Std. ISO 80\,000-1:2022, 2022.

\bibitem{rsna2023}
{Radiological Society of North America}, \emph{RSNA Reporting Initiative v5.0}, 2023.

\bibitem{liu2024medbench}
M.~Liu, W.~Hu, J.~Ding \emph{et~al.}, ``Medbench: A comprehensive, standardized, and reliable benchmarking system for evaluating chinese medical large language models,'' \emph{Big Data Mining and Analytics}, vol.~7, no.~4, pp. 1116--1128, 2024.

\bibitem{lexa2021artificial}
F.~Lexa and S.~Jha, ``Artificial intelligence for image interpretation: Counterpoint-the radiologist's incremental foe,'' \emph{AJR American Journal of Roentgenology}, vol. 217, no.~3, pp. 558--559, 2021.

\bibitem{marcovici2009re}
P.~Marcovici, ``Re: ``frequency and spectrum of errors in final radiology reports generated with automatic speech recognition technology'','' \emph{Journal of the American College of Radiology}, vol.~6, no.~4, pp. 282--283, 2009.

\bibitem{coppola2021human}
F.~Coppola, L.~Faggioni, M.~Gabelloni \emph{et~al.}, ``Human, all too human? an all-around appraisal of the ``artificial intelligence revolution'' in medical imaging,'' \emph{Frontiers in Psychology}, vol.~12, p. 710982, 2021.

\bibitem{fast2024autonomous}
D.~Fast, L.~Adams, F.~Busch \emph{et~al.}, ``Autonomous medical evaluation for guideline adherence of large language models,'' \emph{npj Digital Medicine}, vol.~7, no.~1, p. 358, 2024.

\bibitem{riccardi2024two}
N.~Riccardi, X.~Yang, and R.~Desai, ``The two word test as a semantic benchmark for large language models,'' \emph{Scientific Reports}, vol.~14, no.~1, p. 21593, 2024.

\bibitem{singhal2025toward}
K.~Singhal, T.~Tu, J.~Gottweis \emph{et~al.}, ``Toward expert-level medical question answering with large language models,'' \emph{Nature Medicine}, 2025.

\bibitem{singhal2023large}
K.~Singhal, S.~Azizi, T.~Tu \emph{et~al.}, ``Large language models encode clinical knowledge,'' \emph{Nature}, vol. 620, no. 7972, pp. 172--180, 2023.

\end{thebibliography}
\bibliographystyle{IEEEtran} 


\end{document}